*Article*

# Removing Human Bottlenecks in Bird Classification Using Camera Trap Images and Deep Learning

Carl Chalmers [1], Paul Fergus [1], Serge Wich[1], Steven N Longmore[1], Naomi Davies Walsh[1], Philip Stephens[2], Chris Sutherland[3], Naomi Matthews[4], Jens Mudde[5] and Amira Nuseibeh[2]

1. Liverpool John Moores University; c.chalmers@ljmu.ac.uk, p.fergus@ljmu.ac.uk,s.wich@ljmu.ac.uk, s.n.longmore@ljmu.ac.uk, n.j.walsh@2022.ljmu.ac.uk
2. Department of Biosciences, Durham University; philip.stephens@durham.ac.uk, amira.nuseibeh@durham.ac.uk
3. School of Mathematics and Statistics, University of St Andrews; css6@st-andrews.ac.uk
4. Chester Zoo; n.matthews@chesterzoo.org
5. Department of Biology, Faculty of Science, Utrecht University; j.mudde@students.uu.nl

\* Correspondence: c.chalmers@ljmu.ac.uk

**Abstract:** Birds are important indicators for monitoring both biodiversity and habitat health; they also play a crucial role in ecosystem management. Decline in bird populations can result in reduced eco-system services, including seed dispersal, pollination and pest control. Accurate and long-term monitoring of birds to identify species of concern while measuring the success of conservation interventions is essential for ecologists. However, monitoring is time consuming, costly and often difficult to manage over long durations and at meaningfully large spatial scales. Technology such as camera traps, acoustic monitors and drones provide methods for non-invasive monitoring. There are two main problems with using camera traps for monitoring: a) cameras generate many images, making it difficult to process and analyse the data in a timely manner; and b) the high proportion of false positives hinders the processing and analysis for reporting. In this paper, we outline an approach for overcoming these issues by utilising deep learning for real-time classification of bird species and automated removal of false positives in camera trap data. Images are classified in real-time using a Faster-RCNN architecture. Images are transmitted over 3/4G cameras and processed using Graphical Processing Units (GPUs) to provide conservationists with key detection metrics therefore removing the requirement for manual observations. Our models achieved an average sensitivity of 88.79%, a specificity of 98.16% and accuracy of 96.71%. This demonstrates the effectiveness of using deep learning for automatic bird monitoring.

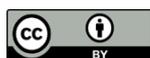

## 1. Introduction

The continuing loss of global biodiversity is a significant concern [1]. Birds are no exception to this phenomenon and a reduction in bird abundance can be a key indicator of adverse environmental change and declines in other species, such as insects [2]. Globally, many bird species have experienced significant population declines [3]. Several factors are implicated in this decline, including climate change [4], agricultural intensification, declining insect abundance and barriers to migration [5] [6]. As a result, organisations and community science programs undertake comprehensive bird monitoring studies to ascertain population sizes while identifying the adverse effects of both natural and anthropogenic events [7].

Impacts on birds tend to vary geographically, and biodiversity in temperate and artic regions is considered to be particularly vulnerable [8]. The long-term monitoring of populations is a critical component of successful conservation [9]. Given that many of the key drivers of decline operate at large scales, monitoring needs to take place over large geographical scales and for long durations. This process is often difficult, resource intensive and costly [10]. In addition, it is estimated that over 85% of species-level information is provided by citizen science projects [11]. This introduces additional challenges due to variance in observer's expertise and biases within monitoring protocols. This can result in both misclassification and inaccurate reporting [12] [13]. As such, conservationists have resorted to the deployment of scalable and cost-effective monitoring technology to improve the modelling of bird populations.

Monitoring technologies such as camera traps and acoustic sensors enable the collection of important ecological data on abundance, distribution, and animal behaviour within ecosystems; this information can be used to stimulate and inform conservation strategies. Camera traps have seen increased usage in bird monitoring and can be complementary to other techniques, such as point counts [13] or acoustic monitoring. Camera traps provide multiple benefits, including unobtrusiveness, low cost, and the ability to conduct studies over large geographical scales and long observation periods. Figure 1 shows an example camera trap image (*Passer domesticus*) from the dataset used in this paper.

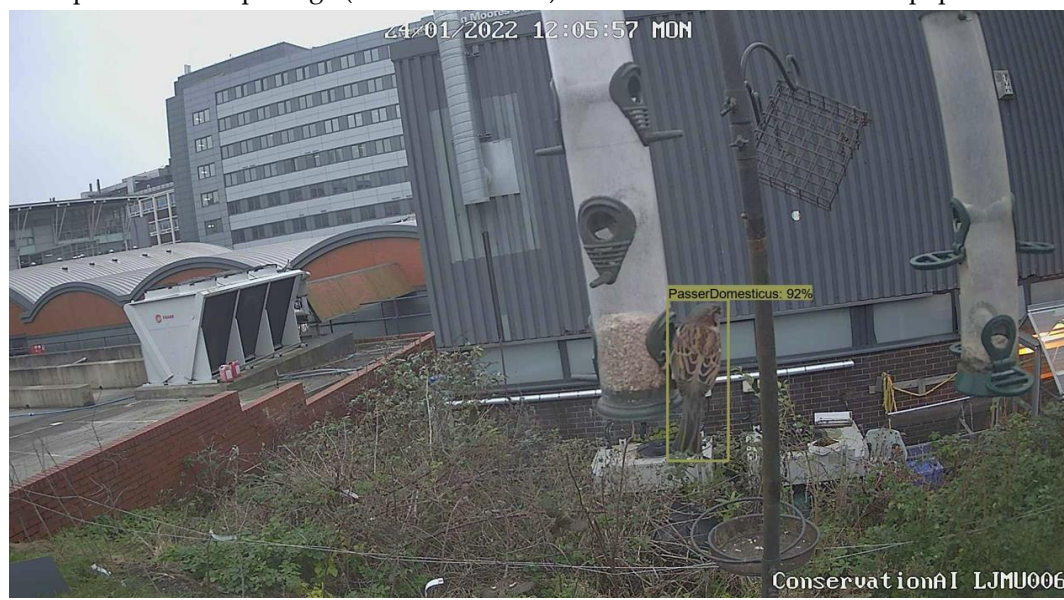

**Figure 1.** Camera trap image of a *Passer domesticus* visiting one of the bird feeders used in the study.

Advancements in camera trap technology, such as improved detection capabilities, extended battery life and wireless communication over 3/4G, have further extended their utility in ecology and conservation. However, their usage has introduced new challenges, including data management and storage, and the manual processing of large quantities of image data [14]. The number of camera traps used in a study can vary greatly, with the average deployment being 78 cameras [15]. As a result, some studies generate millions of images [16], causing significant bottlenecks in data processing and delaying or preventing the completion of studies [17]. The quantity of data collected through camera trap deployments is now one of the main limitations in monitoring studies. The data acquired, and its associated management, analysis and cataloguing requires significant resources to complete. These tasks often lag behind data acquisition, resulting in unused data, and reducing both the accuracy and effectiveness of surveys [18].

Here, we aim to address major challenges of camera trap data processing, through an automated object detection and classification pipeline capable of identifying and cataloguing a variety of different bird species using camera trap data. It is hoped that this approach will support large scale and long-term bird surveys. The technology used in the approach aims to broaden observer participation by allowing citizen scientists and other community groups to participate in biodiversity monitoring. The current version of the pipeline can classify 10 different bird species using deep learning, although additional species could be included following the acquisition of adequate training data. The model is hosted on Conservation AI, which is a free platform that aims to facilitate the processing, management and storage of ecological data.

The remainder of the paper is structured as follows. In Section 2, we provide background to current computer vision tools and their associated limitations, including those associated with automatic species classification. Section 3 details the proposed methodology, including the data description, pre-processing steps, hyperparameter selection and metrics used in the evaluation and inferencing of the trained model. The results for both model training and the deployment of the real-time inferencing trial are presented in Section 4. In Section 5, we discuss the results for both the training and model deployment. Section 6 provides a short conclusion, along with considerations for future work.

## 2. Background and Related Work

The development of computer vision tools for conservation applications is challenging and often impeded by several different factors. These include the availability of high-quality data for each species; high quality requires that data contain sufficient variation in both the species and environment (therefore supporting generalisation) and that the tagging process is accurate and robust. A wide variety of approaches utilise Machine Learning (ML) algorithms such as Artificial Neural Networks (ANNs) for species classification within the mammal taxonomy. However, there is limited research within the *Aves* taxonomy which is largely due to the limited utilisation of camera traps for bird surveys [19] and the complexity of species classification using ML approaches. As such the automatic classification of different bird species has received much less interest among ML practitioners and conservationists. This leaves a significant gap in the tools available for processing data acquired from large scale bird surveys. Additionally, it also limits the involvement of non-domain experts in bird monitoring surveys. The remainder of this section provides a discussion of some of the more common systems in operation to-day and highlights their limitations that we aim to address in this study.

*2.1. Current Solutions*

Historically, the identification of animal species within images has been undertaken manually by humans. However, there is now significant interest in fully or semi-automating this process. In existing approaches, researchers utilise both Deep Learning (DL) and non-DL methods to classify different animals from camera trap data. Most approaches simply detect the presence or absence of an animal to remove blank images and reduce the number of images that require manual processing. However, as this approach does not facilitate species-level classification, practitioners are still required to review large amounts of filtered data. In addition, the approach does little to alleviate the amount of data that has to be stored and managed.

The Microsoft Mega Detector (MD) is one ML tool for removing images which contain no animal objects. In [20], a generic detector was trained using a Faster-RCNN to classify animals, people and vehicles within camera trap data. The model uses the InceptionResNetv2 base network and the TensorFlow Object Detection API. The paper does not report any reliable performance metrics; however, the model has been independently assessed using camera trap data [21][22]. The results showed that it was possible to achieve precision=98% and a recall=93% using a Confidence Threshold (CT) of 0.65 for the Animal class. The results showed that by reducing the CT to 0.1 increased the recall for the Animal class to 97%. However, this decreased precision to 95% therefore reducing the number of false positives. The researchers noted that one of the biggest limitations of MD was its inability to classify individual species. While the MD is a useful tool, it has little benefit for bird surveys.

Other platforms such as Wildlife Insights (WI) do perform species-level classifications of mammals [23]. WI uses DL to detect 800 different animal species using the EfficientNet Architecture. The results show that the model has recall=54% using CT 0.65 across all species indicating a high false negative rate [21]. As a result, low recall values require expert review, leading to a process that is only semi-automated. WI provides a comprehensive platform for processing camera trap data. However, like the MD it has limited utility for bird surveys. In addition, the platform does not facilitate the processing of real-time camera trap data; manual retrieval of data from the SD cards, processing the acquired data and manually uploading it to the WI platform constitute a significant overhead.

Several DL architectures are proposed in [23] to identify the presence or absence of a bird within images. The models were trained with 1200 images with a 125x125 pixel resolution, far lower than the resolution used in camera traps. The paper reported that the best performing model achieved an accuracy of 95.5% after training the model over 60 epochs. No further evaluation metrics were reported while no CTs were used to determine the precision and recall. As such a full assessment of the model's performance is lacking, limiting real-world deployment. Researchers in [24] proposed an approach that utilised sub-category recognition to derive local image features which are later used for classification. The evaluation was performed on the CUB-200-2011 dataset [25]. While the results showed an improvement over existing approaches, only accuracy was reported, and a full evaluation of the model and its inference performance is still required.

Other non-DL approaches have been used for bird classification. In [26] a Histogram of Oriented Gradients (HOG) was combined with Center-Symmetric Local Binary Pattern (CS-LBP) to generate a feature set, which was later classified using a Support Vector Machine (SVM). The classifier was trained to identify a single class (Crow) using a fixed resolution of 48x64 pixels, again far lower than the standard camera trap resolution. The paper reported an accuracy of 87% but no other metrics were provided. In addition, the absence of any additional classes means that the model has limited utility in bird surveys.

*2.2. Limitations*

The literature covers a wide variety of solutions for the automatic removal of blank images. However, very few aim to perform individual species classification. This is particularly true within the *Aves* taxonomy. Where solutions do exist, they are often impractical for real-world inference and do not consider the technical aspects of camera trap technology or the variety of environments in which they are deployed.

Convolutional Neural Network (CNN) approaches require a large corpus of high-quality labelled data that is reflective of the environment (or range of environments) in which the model is deployed. Most of the existing bird datasets used for modelling contain low resolution images or cropped segments of the individual bird. This acts as a denoiser which can often attain high levels of performance during training but perform inadequately during real-world inference. The typical resolution for most camera traps is 1024 x 768. This means that existing models would experience significant drops in performance using larger resolution images as they are compressed or rescaled by the network. This results in extreme distortion of key features that describe the animal in question. Existing research focuses on the identification of a limited number of classes. This approach further reduces the practicality of existing models as many species will go undetected or will likely be misclassified as an alternate species.

Another major challenge to consider is the deployment and automated inference of camera trap data. Large camera trap deployments generate significant amounts of data. This often impedes the study and prevents timely data analysis. The deployment of ML models, either centrally or on the edge, has not been sufficiently reported in the literature.

The centralisation of inferencing will require the use of in field communication such as Global System for Mobile Communications (GSM). While centralising models for inference has many benefits, it requires enhanced hardware to run complex model architectures. Many environments also lack access to GSM. In the remainder of this paper, we will discuss these limitations further and provide a first-step approach for the automatic classification of bird species using DL.

## 3. Materials and Methods

Here, we describe the dataset used in the study, along with the modelling approach taken and the metrics used to evaluate the trained model. We also discuss the data tagging and pre-processing stages, including the proposed and associated technologies.

*3.1. Data Collection and Description*

The image dataset contains images of 10 distinct bird species found in Europe (*Carduelis carduelis*, *Chloris chloris*, *Columba palumbus*, *Corvus corone*, *Cyanistes caeruleus*, *Erithacus rubecula*, *Fringilla coelebs*, *Garrulus glandarius*, *Passer domesticus* and *Pica pica*) which were obtained through the iNaturalist website[1] and private camera deployments.

In total, the dataset contained 32,982 image files. The mean resolution for the dataset was 972 x 769 pixels, conforming to the input resolution of most camera trap hardware. The overall resolution distribution for the entire dataset demonstrates no significant outliers that could negatively impact the training or inferencing of the model (Figure 2). In this case no images were removed before tagging. Undertaking input resolution analysis of the dataset informs the aspect ratio coefficient used in the hyperparameter configuration for training the model.

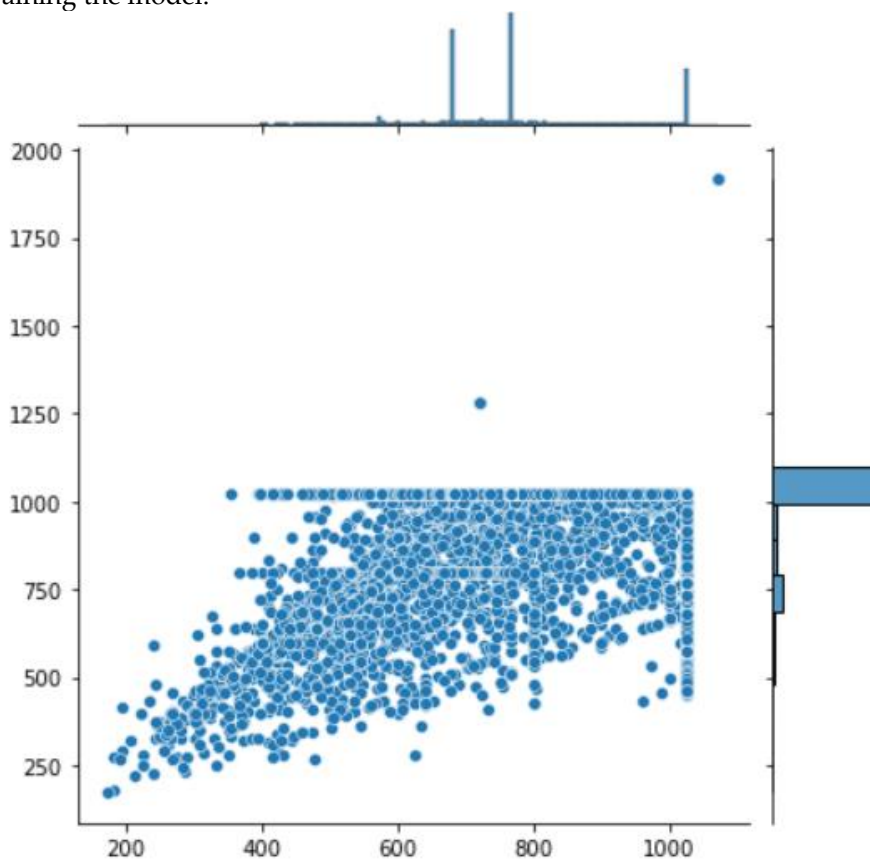

**Figure 2.** Image resolution distribution for the dataset.

---

[1] https://www.inaturalist.org/

*3.2. Data Pre-processing*

The tagging of the data was undertaken using the Conservation AI tagging site. Bounding boxes are placed around objects to identify regions of interest. All tagged regions in each image were exported as Extensible Markup Language (XML) in the TensorFlow Pascal VOC format. Images considered to be poor quality or unsuitable for model training were labelled as no good and systematically removed from the final dataset. The number of images per species varied from ~3,000 to ~3,750, constituting a slight class imbalance (Figure 3). The total number of tagged objects equated to 34,970.

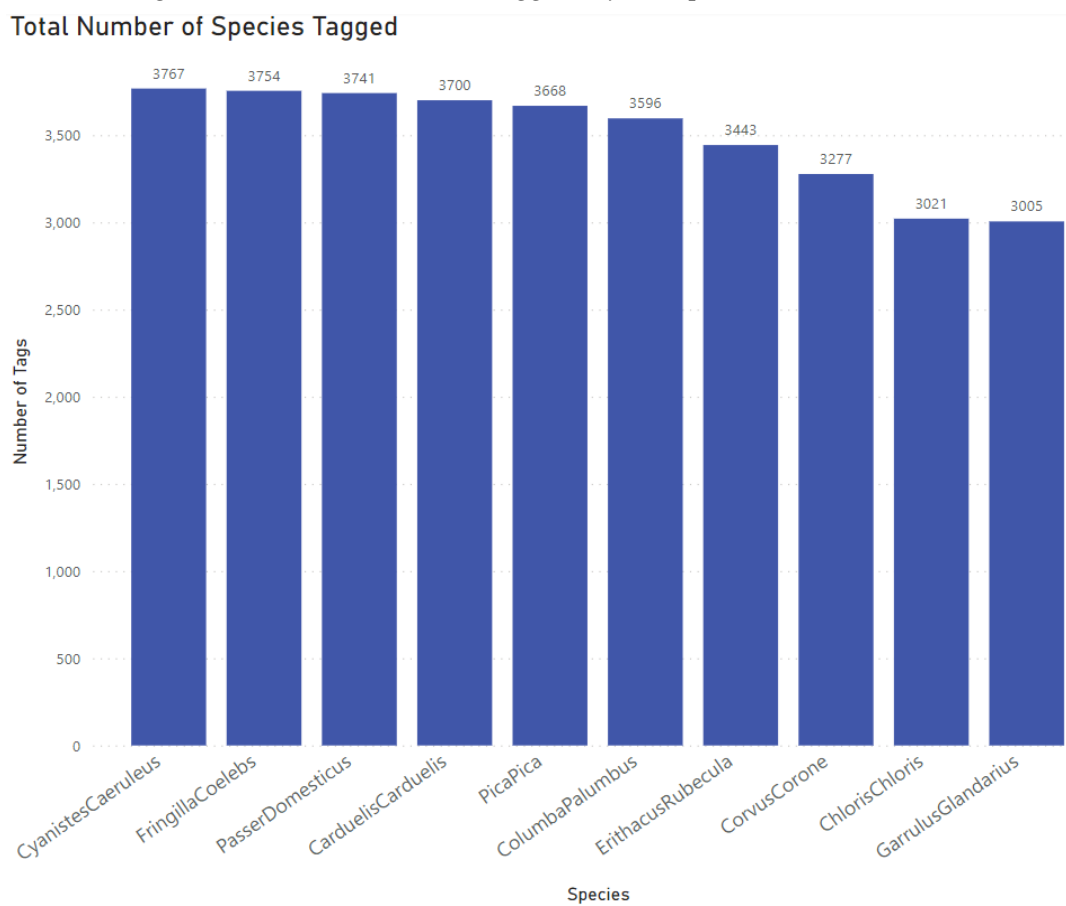

**Figure 3.** Class distribution of tagged data.

Using a python script, the images are randomly divided into 90% train and 10% validation based on the tagged labels. Using TensorFlow 2.2 and Pillow, the XML and associated images are converted into TFRecords (format for storing a sequence of binary records) for model training.

*3.3. Model Selection*

The Faster-RCNN is used to undertake both classification and object detection on images that contain the 10 different bird species. The Faster-RCNN performs object detection in two distinct stages [27]. The first stage uses a region proposal network to identify and extract regions of interest. This allows the model to estimate bounding box locations. The second stage conducts further processing to adjust the localisation of the bounding box by minimising the selected loss function and undertakes classification. Both the region proposal and object detection tasks are undertaken using the same CNN. Figure 4 shows the Faster-RCNN architecture. For a complete description of the network, please refer to [28].

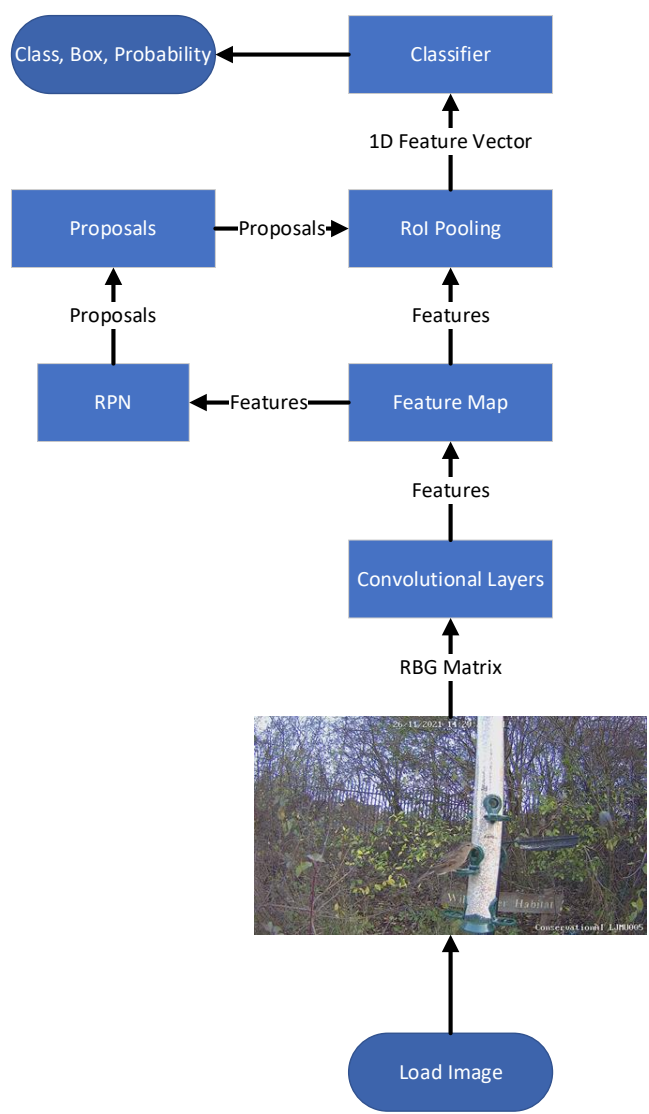

**Figure 4.** Faster RCNN architecture showing the CNN layers, Feature Map, RPN, RoI Pooling and fully connected layers.

*3.4. Transfer Learning*

Transfer learning facilitates the adaptation of a pre-trained model and fine tunes its learned parameters to support new objects of interest. This is an important technique because training CNNs on small datasets leads to poor performance due to low variance and reduced feature representation. Therefore, building on a much larger robust model that has been trained using a large corpus of images helps to counteract the issues associated with the low sample size used in this study. Transfer learning significantly reduces the amount of data required to train a model while maintaining a sufficient level of accuracy. This is an important aspect of this work that enables the development of such solutions to become attainable to a broad range of users with limited data, hardware, and financial support. The base model adopted for the transfer learning task is the Faster-RCNN Resnet 101 model, which has been pre-trained on the Common Objects in Context (COCO) dataset [29]. COCO is a large object detection dataset containing 330 thousand images and 1.5 million object instances.

*3.5. Modelling*

Model training was conducted on an HP ProLiant ML 350 Gen. 9 server with 2x Intel Xeon E5-2640 v4 series processors and 768 GB of RAM. An additional GPU stack comprising 6 Nvidia Quadro RTX8000 graphics cards with a combined total of 288 GB of DDR5 RAM was installed. TensorFlow 2.2.0, CUDA 10.2 and cuDNN version 7.6 formed the software aspects of the training pipeline. In the pipeline.config file used by TensorFlow the following training parameters were set:

- To maintain both the resolution and aspect ratio resizer, minimum and maximum coefficients are set to 1024 x 1024 pixels, respectively. This minimises the scaling effect on the acquired data.
- The default setting for the feature extractor coefficient is retrained to provide a standard 16-pixel stride length to retain the model features and improve training time.
- The batch size coefficient is set to thirty-two, providing a balanced weight update withoutexceeding GPU memory limits.
- The learning rate is set to 0.0004 to prevent large variations in response to the error.

To improve the generalisation of the model and prevent overfitting the following augmentation techniques where applied:

- Random_adjust_hue which adjusts the hue of an image using a random factor.
- Random_adjust_contrast which adjusts the contrast of an image by a random factor.
- Random_adjust_saturation which adjusts the saturation of an image by a random factor.
- Random_square_crop_by_scale which was set with a scale_min of 0.6 and a scale_max of 1.3.

The Adam optimiser is implemented in ResNet 101 to minimise the loss function [30]. Unlike other optimisers such as Stochastic Gradient Decent (SGD), which maintains a single learning rate (alpha) throughout the entire training session, Adam calculates the moving average of the gradient $m_t$/ squared gradients $v_t$ based on the current gradient value $g$, and the parameters $\beta_1$ and $\beta_2$ to dynamically adjust the learning rate. Adam is described in [31] and is defined as:

$$m_t = \beta_1 m_t - 1 + (1 - \beta_1)g_t \tag{1}$$

$$v_t = \beta_2 v_t - 1 + (1 - \beta_2)g_t^2$$

where $m_t$ and $v_t$ are the estimates of the first and second moments of the gradient, respectively. Both $m_t$ and $v_t$ are initialised with 0. Biases are corrected by computing the first and second moment estimates [31] defined as:

$$\hat{m}_t = \frac{m_t}{1 - \beta_1^t} \tag{2}$$

$$\hat{v}_t = \frac{v_t}{1 - \beta_2^t}$$

Parameters are updated using the Adam update rule:

$$\theta_{t+1} = \theta_t - \frac{n}{\sqrt{\hat{v}_t} + \epsilon} \hat{m}_t. \tag{3}$$

The Rectified Linear Unit (ReLU) activation function was adopted during training to provide improvements over other functions such as sigmoid or hyperbolic tangent (tanh) activations that suffer from saturation changes around the mid-point of their input which reduces the amount of available tuning. The use of these in deep multi-layered networks results in ineffective training caused by a vanishing gradient [32]. ReLU, as defined as [33]:

$$g(x) = max(0, x) \qquad (4)$$

*3.6. Model Inferencing*

The trained model was frozen and hosted using TensorFlow 2.5 and served through a public facing website developed by the authors. CUDA 11.2 and cuDNN 7.6.5 enable the GPU accelerated inferencing aspect of the pipeline. Inferencing is undertaken on a custom-built server containing an Intel Xeon E5-1630v3 CPU, 256GB of RAM and a NVidia Quadro RTX 8000 GPU. A NVidia Triton server 22.08 running in Docker with a Windows Subsystem for Linux 2 (WSL2) provided the software stack for inferencing the saved model. Due to the high specification of the GPU and the complexity of the model architecture no model optimisation such as quantization or pruning was used.

A combination of both WIFI and 3/4G cellular cameras was deployed near a set of eight birdfeeders in the UK and the Netherlands as previously shown in figure 1. The camera resolution was configured to 1920 x 1072 pixels with a Dots Per Inch (dpi) of 96. This configuration closely matches the aspect ratio resizer set in the pipeline.config used during training. The IR sensor was set to medium sensitivity which when triggered automatically uploaded the acquired image to the platform using the Simple Mail Transfer Protocol (SMTP) for classification. Each camera was powered using a lithium battery and recharged using solar panels throughout the duration of the study. Figure 5 shows one of the cameras used in the study.

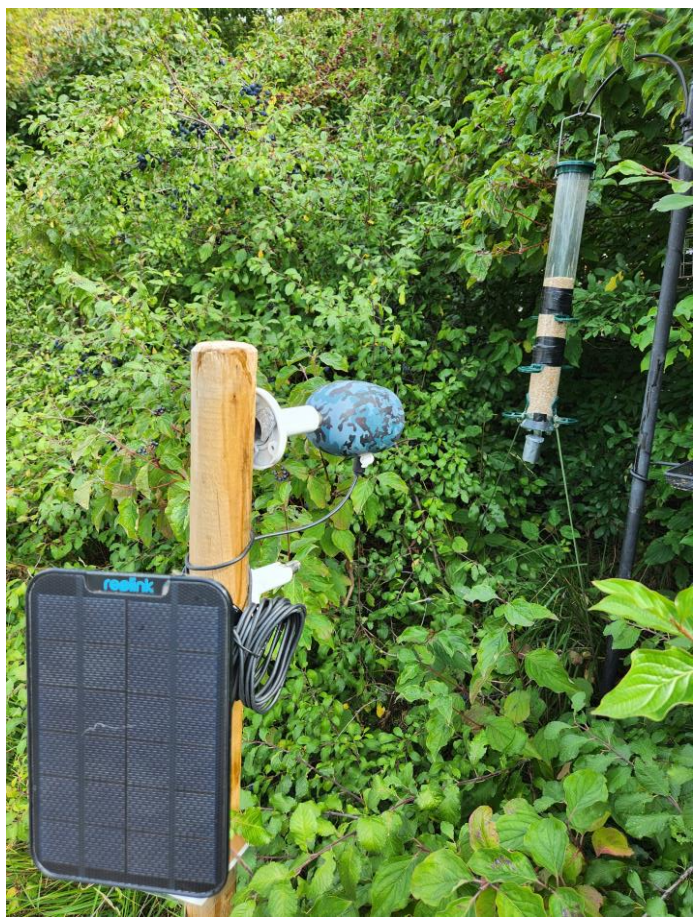

**Figure 5.** 3/4G camera trap used for real-time inference. The camera was deployed near a bird feeder. Throughout the study the camera trap was continuously charged using a solar panel.

The end-to-end inferencing pipeline (Figure 6) starts with the sensor and ends with the public facing Conservation AI site[2] (Figure 7). Due to the use of standard protocols, the system can interface with a variety of different cameras for real-time inference. The SMTP server periodically polls the platforms camera trap email address and automatically schedules the images for classification using the Triton Server RESTAPI. Images are processed as they are received from the real-time camera traps. Where in field communication is unavailable, images files can be batch uploaded through the website, desktop client or REST API for offline inferencing.

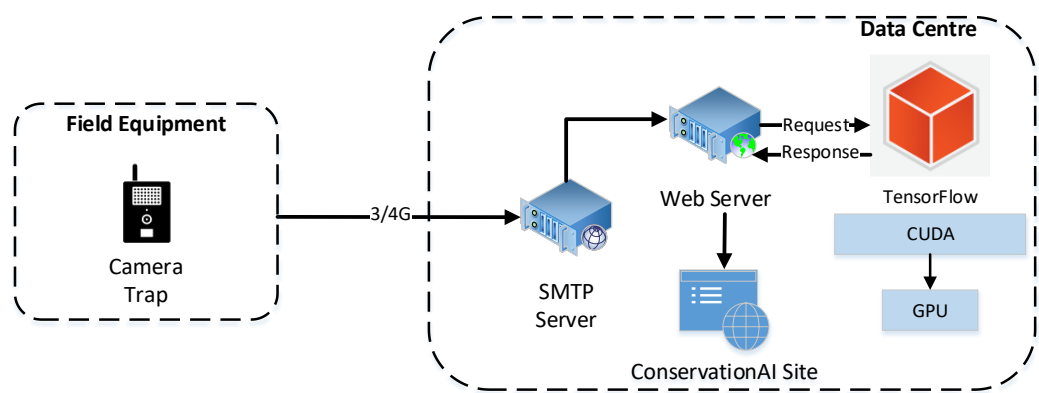

**Figure 6.** End-to-end inferencing pipeline for the Conservation AI platform.

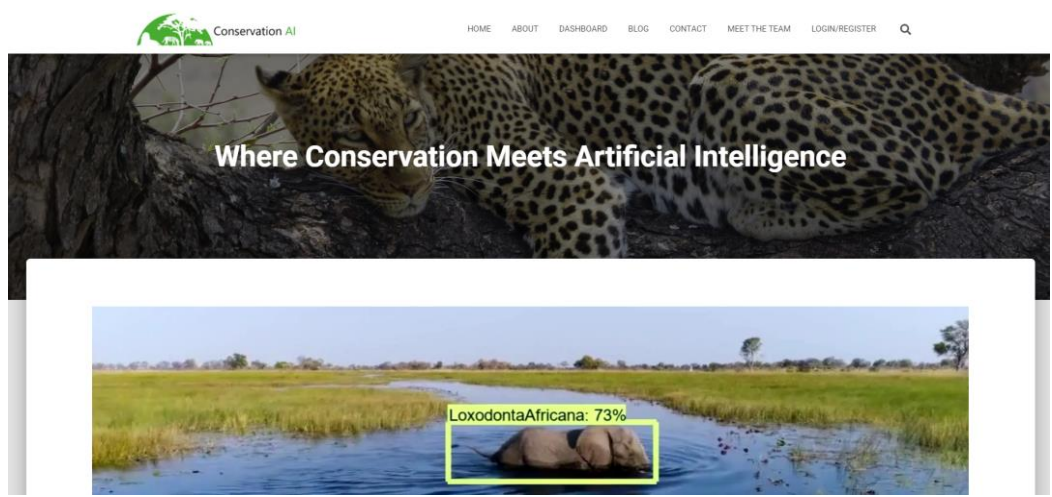

**Figure 7.** Conservation AI platform (https://www.concervationai.co.uk).

The inferencing pipeline (Figure 8) begins with the transmission of acquired image files and associated meta data (Camera ID, image date/time) over 3/4G using SMTP. Images and payload data are automatically extracted from the email body using the imaplib library and sent for classification. Once the image is received by the inference server it is queued for classification. The classified image and associated probability scores are returned and logged to a MySQL database against the user's account, if the probability of a given prediction exceeds 50% confidence. If the probability score is below 50%, the image is classed as a blank (no animal present). Real-time alerts can also be configured, based on the detected species and the classification probability.

---

[2] https://www.conservationai.co.uk/

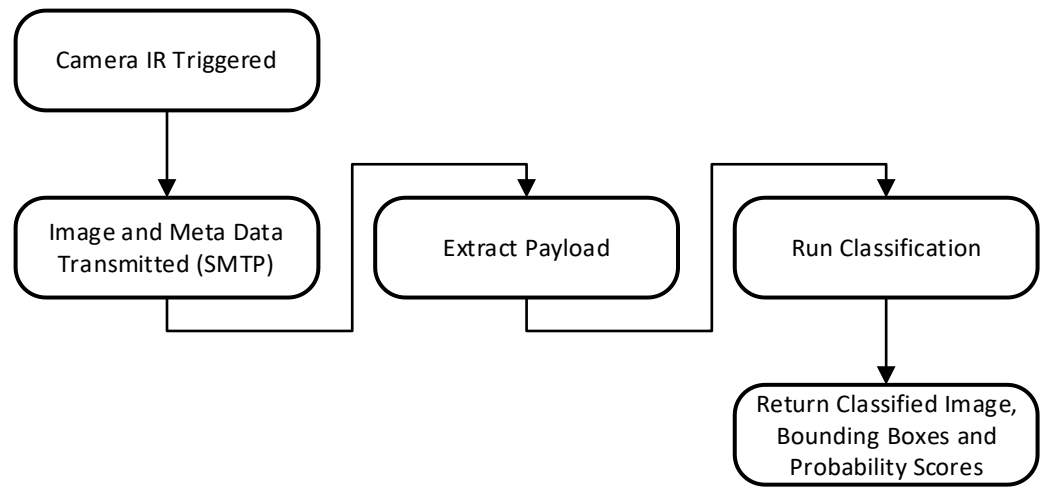

**Figure 8.** Inference processing pipeline.

*3.7. Evaluation Metrics Training*

The performance of the trained model was evaluated using; RPNLoss/objectiveness to measure the model's ability to generate suitable bounding boxes and categorise them as either a foreground or background object; RPNLoss/localisation to measure how well the RPN performs at generating bounding box regressor coordinates for foreground objects; BoxClassifierLoss/classification to measure the output layer/final classifier loss and describe the computed error for prediction and BoxClassifierLoss/localisation to measure the performance of the bounding box regressor. All these measures are combined to produce a TotalLoss metric. These metrics were obtained from TensorBoard 2.6 during training. The validation data was evaluated during training to derive the Mean Average Precision (mAP), which is the standard method for evaluating object detection models. mAP is defined as:

$$mAP = \frac{\sum_{q=1}^{Q} AveP(q)}{Q} \quad (5)$$

Where *Q* is the number of queries in the set and *AveP(q)* is the Average Precision (AP) for a given query q. The mAP is calculated on the bounding box locations for the final two checkpoints. Two Intersection over Union (IOU) thresholds, 0.50 and 0.75, are used to assess the overall performance of the model. The IOU is a useful metric for determining the accuracy of the bounding box location. This is achieved by measuring the percentage ratio of the overlap between the predicted bounding box and the ground truth bounding box as defined as:

$$IoU = \frac{Area\ of\ Overlap}{Area\ of\ Union} \quad (6)$$

A threshold of @0.50 measures the overall detection accuracy of the model while the upper threshold of @0.75 measures localisation accuracy.

*3.8. Evaluation Metrics Inference*

Using the trained model, performance for inference conducted during the trial is evaluated by analysing images transmitted from the real-time cameras. It is important to undertake post-training analysis, as model performance can significantly differ once they are deployed to work on real-world tasks. Precision, Sensitivity (recall), Specificity, F-1 Score and Accuracy are used to evaluate the model's performance on inference data. True Positive (TP), False Positive (FP), True Negative (TN) and False Negative (FN) are used to derive the different metrics. Precision is defined as:

$$Precision = \frac{TP}{TP + FP} \quad (7)$$

Precision is used to ascertain the model's ability to predict true positive detections. In the context of this study, it tells us that the animal detected and classified matched the ground truth while identifying any misclassifications in the image. Recall is defined as:

$$Recall = \frac{TP}{TP + FN} \quad (8)$$

Recall measures the proportion of true positives correctly identified by the model. In this context of object detection, it measures how well the model detected and correctly matched the ground truth labels. Recall also allows us to determine the number of false positives. F-1 Score is defined as:

$$F - 1\ Score = 2 * \frac{Precision * Sensitivity\ (Recall)}{Precision + Sensitivity\ (Recall)} \quad (9)$$

F-1 Score provides the harmonic means of both the precision and recall. A high F-1 Score means that the model has both high precision and recall. In the context of object detection, a high recall means that the model can accurately classify and localise objects in an image. Lastly Accuracy is defined as:

$$Accuracy = \frac{TP + TN}{TP + TN + FP + FN} \quad (10)$$

Accuracy provides an overall assessment of the model's ability to detect and correctly classify objects within the image, although this has limited bearing in unbalanced datasets and should therefore always be interpreted alongside other metrics defined in this section.

## 4. Results

We present the results in two parts: firstly, the training of the Faster-RCNN using the model configuration outlined in the methodology section; secondly, the inference results which demonstrate the model's performance in a real-world setting, using the 3/4G real-time cameras.

### 4.1. Training Results for Europe Birds Model

Using the tagged dataset, the data was randomly split into a training set, comprising 90% of the data (29683 tags), and a validation set of 10% (3298 tags) to train and fine tune the model. The model was trained over 58 epochs (30000 steps) using a batch size of 32 to determine the best fit. During training, there was no overlap between the training and validation loss, indicating that no overfitting occurred.

4.1.1 RPN and Classifier Loss for Training split

The model can detect candidate regions of interest with a high degree of accuracy (0.01325) while the RPN can sufficiently detect the localisation of an object (0.06868) (Table 1). This is encouraging given the size of the tagged objects in relation to the background noise. The classification loss indicates that while the model can effectively localise an object, it is less accurate at correctly predicting the class (0.2977). The results of the box classifier localisation (0.05407) are comparable to the RPN results, affirming that the model can accurately identify candidate labels. The total loss (0.4337) combines the RPN and box classification values. The result demonstrates that the model's predictions closely match the ground truth labels during training. While a total loss of 0.4337 is good for computer vision models, the model evidently found it difficult to classify some species of birds.

**Table 1.** TensorBoard Results for RPN and Classifier Loss on Training Split.

| Metric | Value | Step | Support |
| --- | --- | --- | --- |
| RPNLoss/objectness | 0.01325 | 30000 | 29683 |
| RPNLoss/localization | 0.06868 | 30000 | 29683 |
| BoxClassifierLoss/classification | 0.2977 | 30000 | 29683 |
| BoxClassifierLoss/localisation | 0.05407 | 30000 | 29683 |
| Total Loss | 0.4337 | 30000 | 29683 |

4.1.2 RPN and Classifier Loss for the Validation Split

The validation results are similar to the training results (Table 2). The loss for the RPN objectness (0.01212) is slightly lower than the training set, and the same is true for RPN Localisation (0.04031). Losses for the box classifier (0.2977) and box localisation (0.05407) are again slightly lower than the training metrics. The total loss (0.2828) is lower than the total loss for the training set. This suggests that the images contained within the validation split might have been slightly easier to localise and classify or that the training split contained greater levels of variance. There is no evidence that the model overfitted during training (the training loss does not increase while the validation continues to fall), although further training could have been undertaken to reduce the training loss further.

**Table 2.** TensorBoard Results for RPN and Classifier Loss on Validation Split.

| Metric | Value | Step | Support |
| --- | --- | --- | --- |
| RPNLoss/objectness | 0.01212 | 30000 | 3298 |
| RPNLoss/localization | 0.04031 | 30000 | 3298 |
| BoxClassifierLoss/classification | 0.2496 | 30000 | 3298 |
| BoxClassifierLoss/localisation | 0.0121 | 30000 | 3298 |
| Total Loss | 0.2828 | 30000 | 3298 |

4.1.3 Precision and Recall Results for the Validation Split

Across all classes, the mAP (0.7856) shows that the model was able to detect and correctly identify the class (TP) on most occasions, with only 21.44% being false positives (FP) (Table 3). The mAP(Large), calculated on a subset of the dataset where the object was considered to be large, achieved 0.7902; hence the model was able to correctly identify the class when the object was close to the camera. This process was repeated for both mAP(Medium) and mAP(Small), for which the model's performance dipped to 0.6296 and 0.4629 respectively. This indicates that the model struggles to correctly classify the species when the camera is further away from the object. Both mAP@.50IOU (0.9458) and mAP@.75IOU (0.8992) demonstrate that the model can effectively localise objects in the image and successfully map predicted bounding boxes to the ground truth.

**Table 3.** TensorBoard Results for Precision.

| Metric | Value | Step | Support |
| --- | --- | --- | --- |
| Precision/mAP | 0.7856 | 30000 | 3298 |
| Precision/mAP(Large) | 0.7902 | 30000 | 3298 |
| Precision/mAP(Medium) | 0.6296 | 30000 | 3298 |
| Precision/mAP(Small) | 0.4629 | 30000 | 3298 |
| Precision/mAP@.50IOU | 0.9458 | 30000 | 3298 |
| Precision/mAP@.75IOU | 0.8992 | 30000 | 3298 |

### 4.1.4 Precision and Recall Results for the Validation Split

The model can recall (Recall/AR@1) 81.94% of the objects when considering only the top-ranked detection for each image (Table 4). Similarly, Recall/AR@10 (0.8711) and Recall/AR@100 (0.8721) show the model's performance when considering the top 10 and top 100 detections respectively. The AR@100(Large) can recall 85.57% of the large images, 70.53% of the medium images [AR@100(Medium)] and 55.37% of the small images [AR@100(Small)]. This underlines the model's ability to correctly identify the species if it is closer to the camera, with performance decreasing significantly as objects get smaller (the bird is further away from the camera).

**Table 4.** TensorBoard Results for Recall.

| Metric | Value | Step | Support |
| --- | --- | --- | --- |
| Recall/AR@1 | 0.8194 | 30000 | 3298 |
| Recall/AR@10 | 0.8711 | 30000 | 3298 |
| Recall/AR@100 | 0.8721 | 30000 | 3298 |
| Recall/AR@100(Large) | 0.8759 | 30000 | 3298 |
| Recall/AR@100(Medium) | 0.7053 | 30000 | 3298 |
| Recall/AR@100(Small) | 0.5537 | 30000 | 3298 |

### 4.2. Trial Results for European Birds Model

The trained model was deployed to assess its performance during the trial period. Using eight different cameras and the inferencing pipeline outlined previously in the methodology section, the cameras transmitted a total of 42,973 images between 05/03/2021 and 28/02/2023. During the trial, 14,740 detections were made of the 10 species (Figure 9) and 28,233 blanks were recorded.

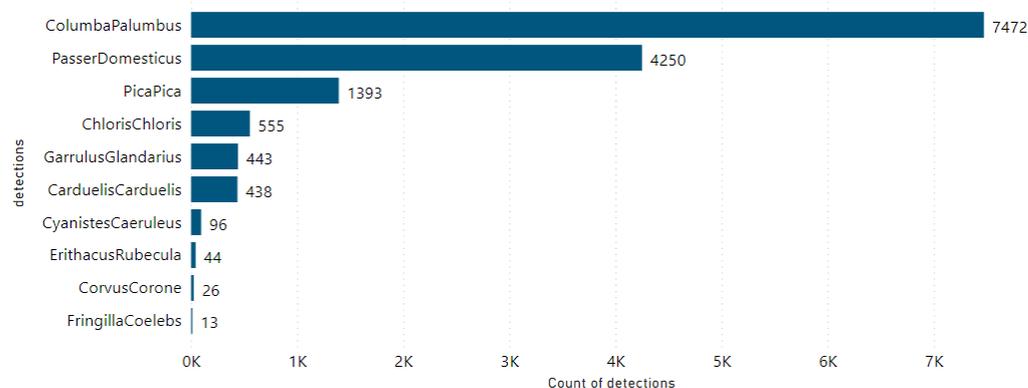

**Figure 9.** Number of detections by species during the study duration.

The 14,740 detections were used to generate ten inference evaluation subsets each containing 25 randomly sampled images from each of the eight classes (excluding the *Corvus corone* and *Fringilla coelebs* classes, as these were insufficiently represented in the data). In addition, 25 blanks were randomly sampled and added to each fold to generate a total of 250 images per fold. A 5% margin of error and a confidence level of 95% was used to generate the inference evaluation dataset. The metrics were calculated for each of the 10 datasets and averaged to produce a final set for the model.

### 4.2.1 Performance Evaluation Results for Inference

The model archived a high level of accuracy across all classes, with an average range between 95.19% and 98.28%, and an overall accuracy of 96.71 across the 10 folds (Table 5). However, the model's precision scores are more diverse. The precision ranges between 35.02% and 100% with the *Garrulus glandarius* class having the lowest value. Overall, the model's average precision (75.32) is good for an object detection model. However, this indicates that there are issues with several classes (*Carduelis carduelis*, *Chloris chloris* and *Garrulus glandarius*). This requires further investigation of the training data and tagging process for these classes. The sensitivity of the model is high across all classes (between 76.67% and 95.24%). The worst performing classes for sensitivity were *Cyanistes caeruleus* (76.67%) and *Erithacus rubecula* (85.73%), which need further investigation to ascertain the reason for lower scores. However, it is likely a result of the model's dependence on the colour of these two species and the data augmentation techniques applied during training. The average sensitivity of 88.79% across all folds indicates that the model can correctly identify a large proportion of actual positive cases. Specificity for the model is high across all classes (95.47% to 100%). The species with the lowest specificity was *Chloris chloris*; the highest was *Erithacus rubecula*. The model's average specificity across all folds was 98.16% indicating that the model was able to correctly identify most of the negative cases (distinguishing correctly between images that have an object in and images that are blank). The model's F1-score varied across all the classes (between 47.61% and 93.49%). The species with the highest F1-score was *Columba pa-lumbus* (92.51%); the lowest was *Garrulus glandarius* (47.61%). The average F1-score across the folds was 77.10%, which requires further investigation.

**Table 5.** Inference Performance Metrics for Camera Deployment.

|  | Fold1 | Fold2 | Fold3 | Fold4 | Fold5 | Fold6 | Fold7 | Fold8 | Fold9 | Fold10 | Average |
|---|---|---|---|---|---|---|---|---|---|---|---|
| *Carduelis carduelis* | | | | | | | | | | | |
| Accuracy | 91.58% | 96.39% | 96.39% | 97.81% | 96.35% | 97.80% | 96.00% | 99.37% | 98.82% | 98.19% | 96.87% |
| Precision | 85.00% | 93.75% | 93.75% | 70.00% | 30.00% | 33.33% | 12.50% | 50.00% | 60.00% | 25.00% | 55.33% |
| Sensitivity | 56.67% | 71.43% | 71.43% | 87.50% | 100.00% | 100.00% | 100.00% | 100.00% | 100.00% | 100.00% | 88.70% |
| Specificity | 98.13% | 99.42% | 99.42% | 98.29% | 96.30% | 97.78% | 95.98% | 99.36% | 98.80% | 98.18% | 98.17% |
| F1-Score | 68.00% | 81.08% | 81.08% | 77.78% | 100.00% | 50.00% | 22.22% | 66.67% | 75.00% | 40.00% | 66.18% |
| *Chloris chloris* | | | | | | | | | | | |
| Accuracy | 92.55% | 95.90% | 93.58% | 92.75% | 98.40% | 94.68% | 97.11% | 96.91% | 95.45% | 97.02% | 95.44% |
| Precision | 36.84% | 52.94% | 42.11% | 26.32% | 80.00% | 44.44% | 54.55% | 61.54% | 46.15% | 50.00% | 49.49% |
| Sensitivity | 77.78% | 100.00% | 88.89% | 100.00% | 100.00% | 100.00% | 100.00% | 100.00% | 85.71% | 100.00% | 95.24% |
| Specificity | 93.30% | 95.70% | 93.82% | 92.55% | 98.30% | 94.44% | 97.01% | 96.75% | 95.86% | 96.93% | 95.47% |
| F1-Score | 50.00% | 69.23% | 57.14% | 41.67% | 88.89% | 61.54% | 70.59% | 76.19% | 60.00% | 66.67% | 64.19% |
| *Columba pa-lumbus* | | | | | | | | | | | |
| Accuracy | 93.55% | 96.39% | 95.63% | 91.33% | 97.37% | 97.80% | 94.38% | 94.58% | 94.38% | 96.45% | 95.19% |
| Precision | 97.96% | 96.36% | 95.74% | 93.22% | 98.61% | 96.55% | 96.72% | 92.31% | 92.06% | 95.16% | 95.47% |
| Sensitivity | 81.36% | 91.38% | 88.24% | 80.88% | 94.67% | 96.55% | 88.06% | 90.57% | 92.06% | 95.16% | 89.89% |
| Specificity | 99.21% | 98.53% | 98.48% | 96.88% | 99.13% | 98.39% | 98.20% | 96.46% | 95.65% | 97.20% | 97.81% |
| F1-Score | 88.89% | 93.81% | 91.84% | 86.61% | 96.60% | 96.55% | 92.19% | 91.43% | 92.06% | 95.16% | 92.51% |
| *Cyanistes caeruleus* | | | | | | | | | | | |
| Accuracy | 91.58% | 98.42% | 95.63% | 95.72% | 96.35% | 97.27% | 97.27% | 94.92% | 96.91% | 97.11% | 96.12% |
| Precision | 65.52% | 92.00% | 95.83% | 100.00% | 100.00% | 100.00% | 100.00% | 94.74% | 100.00% | 92.31% | 94.04% |
| Sensitivity | 76.00% | 95.83% | 76.67% | 70.37% | 73.08% | 73.08% | 78.26% | 69.23% | 79.17% | 75.00% | 76.67% |
| Specificity | 93.94% | 98.80% | 99.35% | 100.00% | 100.00% | 100.00% | 100.00% | 99.34% | 100.00% | 99.36% | 99.08% |
| F1-Score | 70.37% | 93.88% | 85.19% | 82.61% | 84.44% | 84.44% | 87.80% | 80.00% | 88.37% | 82.76% | 83.99% |
| *Erithacus rubecula* | | | | | | | | | | | |
| Accuracy | 94.57% | 97.91% | 97.77% | 96.24% | 98.93% | 96.74% | 97.67% | 98.74% | 97.67% | 96.45% | 97.27% |
| Precision | 89.66% | 95.83% | 95.83% | 100.00% | 100.00% | 95.83% | 92.00% | 100.00% | 100.00% | 95.83% | 96.50% |
| Sensitivity | 78.79% | 88.46% | 88.46% | 76.67% | 92.31% | 82.14% | 92.00% | 91.67% | 84.62% | 82.14% | 85.73% |
| Specificity | 98.01% | 99.39% | 99.35% | 100.00% | 100.00% | 99.36% | 98.64% | 100.00% | 100.00% | 99.29% | 99.40% |
| F1-Score | 83.87% | 92.00% | 92.00% | 86.79% | 96.00% | 88.46% | 92.00% | 95.65% | 91.67% | 88.46% | 90.69% |
| *Garrulus glandarius* | | | | | | | | | | | |
| Accuracy | 93.05% | 97.40% | 97.77% | 97.81% | 97.88% | 98.89% | 100.00% | 94.58% | 98.25% | 99.39% | 97.50% |
| Precision | 7.69% | 37.50% | 33.33% | 20.00% | 33.33% | 33.33% | 100.00% | 10.00% | 25.00% | 50.00% | 35.02% |
| Sensitivity | 50.00% | 100.00% | 100.00% | 100.00% | 100.00% | 100.00% | 100.00% | 100.00% | 100.00% | 100.00% | 95.00% |
| Specificity | 93.51% | 97.35% | 97.74% | 97.80% | 97.86% | 98.88% | 100.00% | 94.55% | 98.24% | 99.39% | 97.53% |
| F1-Score | 13.33% | 54.55% | 50.00% | 33.33% | 50.00% | 50.00% | 100.00% | 18.18% | 40.00% | 66.67% | 47.61% |
| **Blank** | | | | | | | | | | | |
| Accuracy | 97.75% | 98.94% | 98.87% | 98.35% | 98.93% | 99.44% | 97.67% | 98.13% | 97.11% | 97.60% | 98.28% |
| Precision | 100.00% | 100.00% | 100.00% | 100.00% | 100.00% | 100.00% | 100.00% | 100.00% | 100.00% | 100.00% | 100.00% |
| Sensitivity | 83.33% | 92.00% | 91.67% | 88.00% | 92.00% | 96.00% | 84.00% | 88.00% | 80.00% | 84.00% | 87.90% |
| Specificity | 100.00% | 100.00% | 100.00% | 100.00% | 100.00% | 100.00% | 100.00% | 100.00% | 100.00% | 100.00% | 100.00% |
| F1-Score | 90.91% | 95.83% | 95.65% | 93.62% | 95.83% | 97.96% | 91.30% | 93.62% | 88.89% | 91.30% | 93.49% |
| *Passer domesticus* | | | | | | | | | | | |
| Accuracy | 96.13% | 96.89% | 96.15% | 93.72% | 94.87% | 95.19% | 96.00% | 96.91% | 96.00% | 95.88% | 95.77% |
| Precision | 94.44% | 97.22% | 88.57% | 87.50% | 85.71% | 97.56% | 92.11% | 97.06% | 95.12% | 91.43% | 92.67% |
| Sensitivity | 87.18% | 87.50% | 91.18% | 87.50% | 80.00% | 83.33% | 92.11% | 89.19% | 88.64% | 88.89% | 87.55% |
| Specificity | 98.59% | 99.35% | 97.30% | 95.80% | 97.58% | 99.28% | 97.08% | 99.20% | 98.47% | 95.88% | 97.85% |
| F1-Score | 90.67% | 92.11% | 89.86% | 87.50% | 82.76% | 89.89% | 90.91% | 92.96% | 91.76% | 90.14% | 89.85% |
| *Pica pica* | | | | | | | | | | | |
| Accuracy | 96.67% | 97.91% | 98.31% | 95.21% | 98.93% | 99.44% | 97.67% | 98.13% | 97.67% | 98.19% | 97.81% |
| Precision | 25.00% | 50.00% | 80.00% | 38.46% | 77.78% | 85.71% | 50.00% | 75.00% | 75.00% | 57.14% | 61.41% |
| Sensitivity | 100.00% | 75.00% | 88.89% | 83.33% | 100.00% | 100.00% | 100.00% | 60.00% | 60.00% | 100.00% | 86.72% |
| Specificity | 96.63% | 98.40% | 98.82% | 95.60% | 98.89% | 99.42% | 97.62% | 99.35% | 99.35% | 98.15% | 98.22% |
| F1-Score | 40.00% | 60.00% | 84.21% | 52.63% | 87.50% | 92.31% | 66.67% | 66.67% | 66.67% | 72.73% | 68.94% |
| **Overall Model** | | | | | | | | | | | |
| Accuracy | 94.16% | 97.35% | 96.71% | 95.44% | 97.56% | 97.47% | 96.83% | 97.14% | 96.94% | 97.49% | 96.71% |
| Precision | 66.90% | 79.51% | 79.79% | 70.61% | 78.38% | 76.31% | 76.69% | 76.21% | 74.92% | 73.84% | 75.32% |
| Sensitivity | 76.79% | 89.07% | 87.83% | 86.03% | 92.45% | 92.92% | 91.71% | 88.73% | 89.56% | 92.80% | 88.79% |
| Specificity | 96.81% | 98.55% | 98.18% | 97.44% | 98.67% | 98.62% | 98.21% | 98.41% | 98.22% | 98.54% | 98.16% |
| F1-Score | 66.23% | 81.39% | 80.79% | 71.39% | 80.91% | 79.39% | 78.43% | 76.64% | 77.77% | 78.11% | 77.10% |

4.2.2 Confusion Matrix for Inference Data

Sources of confusion among the inference data across all folds (Table 6) largely align with the inference scores in Table 5. The model correctly predicts most of the classes with minimal error; the *Columba palumbus*, *Chloris chloris* and *Passer domesticus* classes performed the best. However, the model does misclassify examples across all the species with *Cyanistes caeruleus* performing the worst. Again, this closely matches the results in Table 5 and is likely due to the model's dependence on colour and the augmentation techniques used in training. These results highlight shortcomings in the model's ability to accurately detect and classify some species which will need to be addressed in future training iterations.

**Table 6.** Confusion Matrix for Inference. The bold diagonal number indicate the TP for each of the classes.

|  | *Carduelis carduelis* | *Chloris chloris* | *Columba palumbus* | *Cyanistes caeruleus* | *Erithacus rubecula* | *Garrulus glandarius* | *Blank* | *Passer domesticus* | *Pica pica* |
|---|---|---|---|---|---|---|---|---|---|
| *Carduelis carduelis* | **63** | 11 | 0 | 8 | 0 | 4 | 0 | 0 | 1 |
| *Chloris chloris* | 1 | **74** | 0 | 3 | 0 | 0 | 0 | 0 | 0 |
| *Columba palumbus* | 4 | 0 | **552** | 0 | 0 | 22 | 0 | 8 | 28 |
| *Cyanistes caeruleus* | 14 | 25 | 5 | **187** | 0 | 4 | 0 | 1 | 5 |
| *Erithacus rubecula* | 1 | 28 | 1 | 0 | **232** | 2 | 0 | 8 | 0 |
| *Garrulus glandarius* | 0 | 0 | 0 | 1 | 0 | **14** | 0 | 0 | 0 |
| *Blank* | 0 | 2 | 9 | 3 | 5 | 0 | **218** | 11 | 0 |
| *Passer domesticus* | 13 | 14 | 7 | 0 | 4 | 11 | 0 | **345** | 0 |
| *Pica pica* | 0 | 0 | 4 | 0 | 0 | 1 | 0 | 0 | **49** |

## 5. Discussion

Bird monitoring provides an important biodiversity indicator but is a complex and time-consuming task, largely accessible only to skilled participants. Coupled with the need to analyse large volumes of data, this impedes the speed at which inferences can be used to inform interventions for the prevention of biodiversity loss. We demonstrate a practical alternative, that uses deep learning and real-time 3/4G cameras for the automatic classification of bird species. The model was trained on 34,970 tags across 10 different bird species, using the Faster-RCNN architecture and transfer learning. The model allowed us to monitor eight different sites in both the UK and Netherlands over a two-year period.

During training, the model attained good results for the RPNLoss/objectness (0.01325) and RPNLoss/localistion (0.04031) for the training split. Both the validation RPNLoss/objectness and RPNLoss/localistion performed equally well, with a loss of 0.01212 and 0.04031 respectively. The BoxClassifierLoss/classification (0.2977) and BoxClassifierLoss/localtistion (0.05407) for the training split performed reasonably well although the higher classification loss indicates that the model struggled to correctly classify all species of birds. This was reflected in validation, during which the BoxClassifierLoss/classification and BoxClassifierLoss/localisation attained a loss of 0.2496 and 0.0121, respectively. By combining the metrics, the model was able to achieve a total loss of 0.4337 (train split) and 0.2828 (validation split), without overfitting. Given both the batch size and the number of tags used during training, the model would have benefited from a larger number of epochs.

The model performed well and attained good precision results on the validation split. The metrics demonstrate that the model can detect and localise objects with a high degree of accuracy. However, the model struggles with smaller and more distant objects, likely characterised by fewer features. Thi is reflected in the recall results for the validation split.

During the trial the model inference performed well achieving an average accuracy (96.71%), precision (75.32%), sensitivity (88.79%), specificity (98.16%) and an F1-score (77.10) across the 10 folds. However, the model struggled with some species such as the *Chloris chloris* and *Carduelis carduelis*, with average precisions of only 49.49% and 55.33% respectively. Although the species was sometimes reasonably prominent within images (e.g., Figure 10), they were often distant from the cameras in the inference dataset. In contrast to other species, they also have a limited number of identifiable features at greater distances. However, the model is capable of detecting complex patterns when the bird is closer to the camera (e.g., Figure 11). It is probable that the model depends largely on the objects colour to identify this class. The type of data augmentation used during training, during which colour was randomly altered to improve generalisation, will have compounded problems when the mode tries to identify this species.

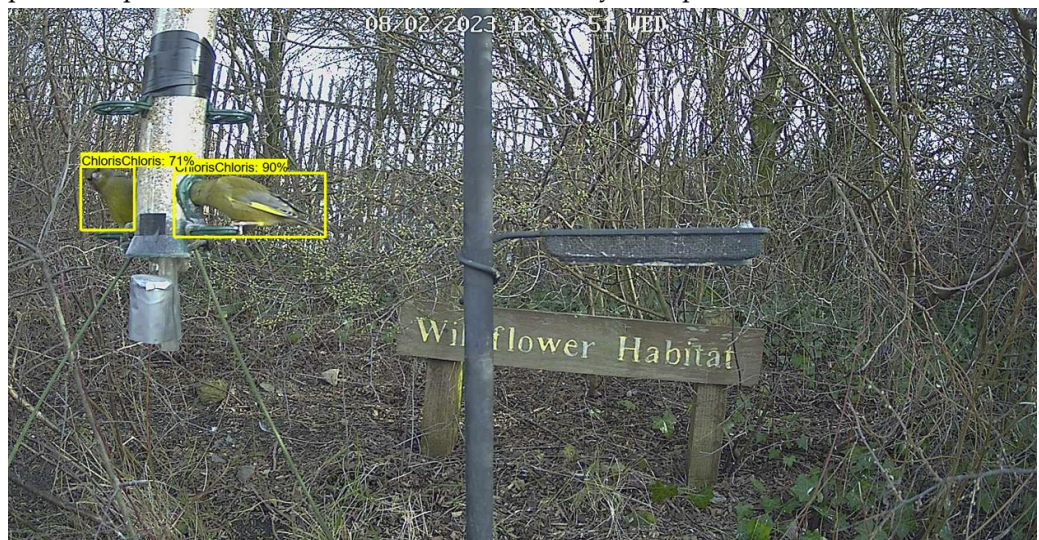

**Figure 10.** Example detections of *Chloris chloris* during the inference trial using one of the 3/4G cameras.

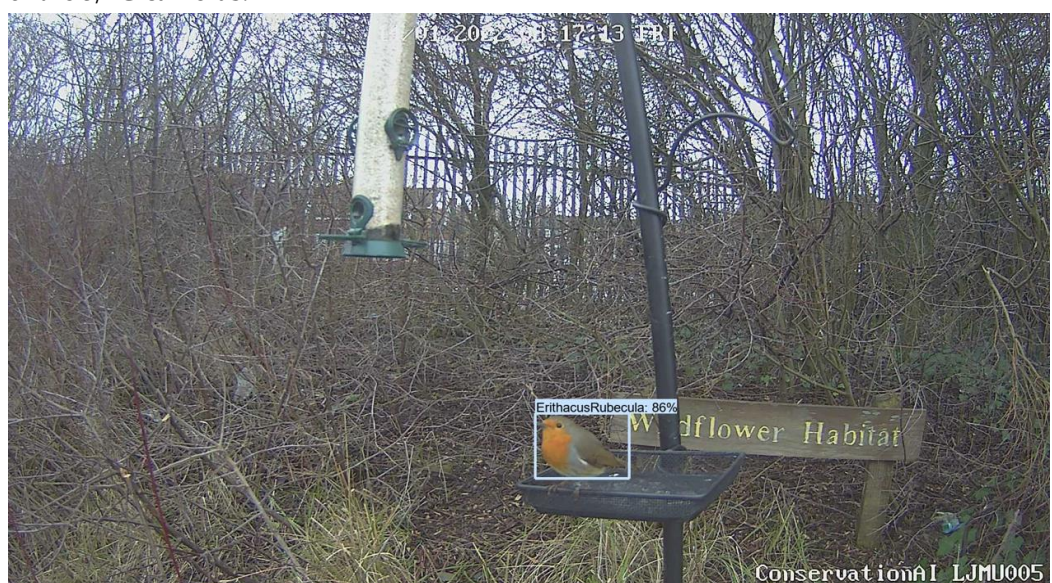

**Figure 11.** Example detection of a *Erithacus rubecula* during the inference trial using one of the 3/4G cameras.

Clearly, the model performs better with larger objects. With an average precision of 95.47%, performance for *Columba palumbus* was much better – likely a result of their large size and greater number of visible features (e.g., Figure 12 and Figure 13). The distance of the object from the camera clearly plays a significant role in the successful classification of a species. This means that careful consideration is needed when deploying cameras for small animals such as birds. Cameras must be deployed as close to the bird feeder as possible to increase the likelihood of a successful classification.

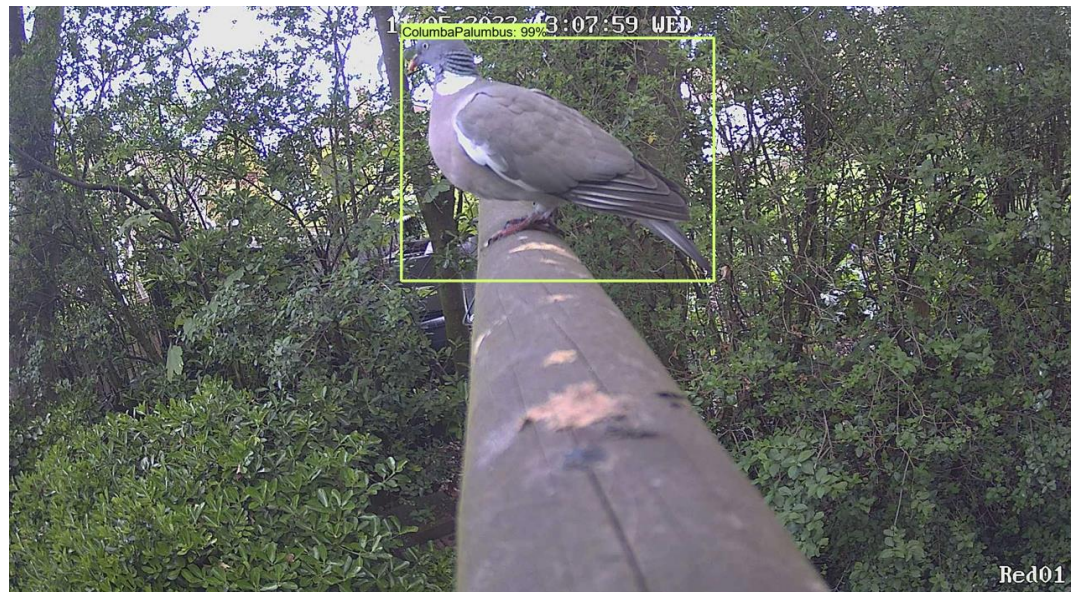

**Figure 12.** Example detections of *Columba palumbus* during the inference trial.

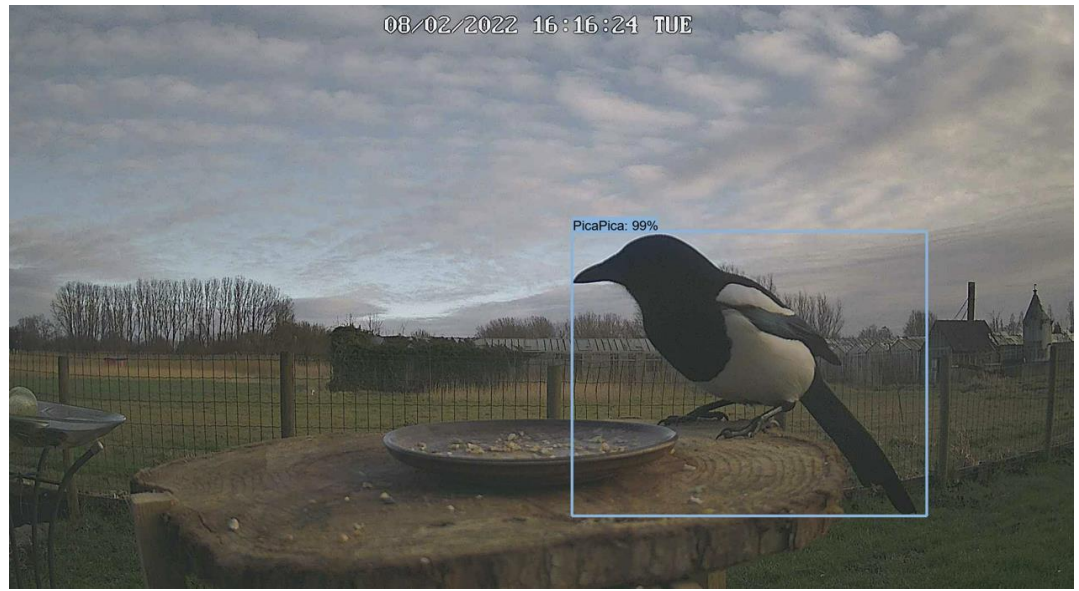

**Figure 13.** Example detections of *Pica pica* during the inference trial.

The inference results demonstrate that the model can detect an object with a high degree of accuracy. Likewise, it is also highly effective at removing images that do not contain a bird. During the trial, the model was able to remove blank images with an average accuracy of 98.28%. Practitioners should exercise caution when using the model for the classification of both *Passer domesticus* and *Columba palumbus*. The model will assign the closet matching class to species the model has not yet been trained on. It was noted during evaluation that the model sometimes mistook the *Passer domesticus* class for *Prunella modularis* (a similar looking species). This is likely due to similarities between the species, occlusion (therefore reducing available features) and incorrect tagging in the training data. This is also true for the for *Columba palumbus* where the model can predict different species within the same genus.

Overall, there has been limited research on the effectiveness of camera traps for monitoring small animals such as birds. The method could also be reduced if the camera fails to trigger, or the speed of the camera results in blurry or distorted images [34]. The inference test set contained several blurry images which negatively affected the results. This resulted from a low trigger speed on the deployed cameras. While this might have limited adverse effects on larger animals, distorted image can remove some of the detailed features of smaller animals which the model relies on for classification.

## 6. Conclusions

We presented an automatic bird classification system capable of identifying ten different species using both deep learning and real-time 3/4G cameras. This provides a possible solution to some of the issues associated with conducting large-scale bird monitoring projects, by reducing the requirements for domain expertise and the manual classification of camera trap data. Removing the bottlenecks in manual camera trap processing will speed up information provision and enable timely interventions for biodiversity. Using a transfer learning approach and the Faster-RCNN architecture, a model was trained and deployed in two countries to ascertain its effectiveness for monitoring bird populations. Images were transmitted in real-time and inferenced on the Conservation AI platform to automatically document the species of birds visiting each location. The results of the trial are encouraging and demonstrate that individual species can be detected with a high degree of accuracy. The model is also effective at accurately removing the large numbers of blank images that are transmitted because of motion detection. This is particularly important in bird studies, in which camera traps need to be set to higher levels of sensitivity due to the small size of the focal objects. By hosting the trained model on the Conservation AI platform, a unified and seamless process capable of automatically documenting bird biodiversity is provided. The solution should be seen as a foundational step that not only benefits domain experts but provides a gateway for anyone with an interest in contributing to the monitoring of bird populations. For example, it would be very easy to offer people the chance to deploy their own cameras in gardens, workspaces, and even communal parks to help broaden the bird monitoring capabilities of organisations like RSPB to better understand the biodiversity health of bird species in countries like the UK.

Our results identify important avenues for further research. While reducing the camera's distance from the bird feeder will likely improve classification results for smaller bird species, alterations in the training data, tagging and hyperparameters are likely to yield greater improvements in accuracy. Removing data augmentation techniques should be trialled to yield significant gains in model performance. In addition, further work is required regarding the optimal camera setup and configuration for computer vision models in bird monitoring. Overall, we feel that the work demonstrates promising results and shows how deep learning and real-time inference can aid in the monitoring of birds, especially in those areas in which large numbers of skilled participants are unavailable.

**Acknowledgments:** The authors would like to thank ReoLink for donating the cameras and solar panels used in this study and Vodafone UK for sponsoring our communication. Finally, the authors would like to thank Rachel Chalmers for tagging all the data used in this study and for contributing so much to the Conservation AI platform.